\documentclass[12pt]{article}
\usepackage[a4paper,margin=1in]{geometry}
\usepackage[T1]{fontenc}
\usepackage[utf8]{inputenc}
\usepackage{mathptmx}
\usepackage{setspace}
\usepackage{amsmath,amssymb,amsfonts,amsthm}
\usepackage{graphicx}
\usepackage{float}
\usepackage{xcolor}
\usepackage[section]{placeins}
\usepackage[colorlinks=true,linkcolor=blue,citecolor=blue,urlcolor=blue]{hyperref}

\newtheorem{proposition}{Proposition}
\newtheorem{remark}{Remark}

\title{What Does Machine Learning Learn from Numerical Calabi--Yau Geometry?\\[0.5em]
{\large A Controlled Study on the Fermat Quintic}}
\author{D Yang Eng}
\date{}

\begin{document}
\doublespacing
\maketitle

\begin{abstract}
We study polynomial regression of numerical metric errors on the Fermat
quintic. A known hypersurface volume-ratio identity gives the exact
reference $C-\log p_4$ relative to the induced Fubini--Study
metric, where $p_4=\sum_i|z_i|^8$ on unit-normalized homogeneous coordinates.
We optimize algebraic metrics at degrees $6,8,10,12$ while preserving finite
Fermat symmetry, then fit their remaining logarithmic volume-form error.
Regression splits and bootstrap draws keep the five roots of each sampled
line together, with all compared teachers evaluated on common clouds.
Warm-start replication correlations exceed $0.9997$ at $k=8,10,12$.
Across degrees, source-derived modes fall short of target-specific fits,
including in a post-hoc quartic check, but outperform random spans of
matched rank in the tested libraries. Two independent random-invariant
$k=10$ starts have residual correlations $0.768$--$0.803$, while their fitted
cubic modes correlate at $0.979$--$0.983$ and retain useful calibrated
cross-start prediction. Their fitted modes agree poorly with those of the
principal warm-start teacher. Adding the phase-sensitive invariant $J_7$
improves prediction beyond the cubic modulus-only library for all seven
warm-start teachers. The gain depends on the library: at $k=12$, a quartic
modulus-only fit explains about 97--98\% of residual variance and the
conditional interval for the added $J_7$ gain includes zero for both replicas
under both measures. The gains therefore establish neither phase
independence nor irreducible phase necessity; fixed-baseline augmentation
preserves this distinction. The fitted components depend
on the numerical protocol; global optimality and a universal finite-degree
correction remain unestablished.
\end{abstract}



\section{Introduction}

Ricci-flat Calabi--Yau metrics are important in complex geometry and string
compactification. Yau's proof of the Calabi conjecture guarantees a unique
Ricci-flat K\"ahler metric in each K\"ahler class on a compact Calabi--Yau
manifold \cite{Yau1978}. The theorem does not provide explicit coordinate
expressions for these metrics. For examples such as the quintic threefold,
no explicit closed-form Ricci-flat metric is currently known
\cite{LeeLukas2026}.

Donaldson's balanced-metric program provides an approximation scheme using
holomorphic sections of positive line bundles
\cite{Donaldson2001,Donaldson2009}. Douglas, Karp, Lukic and Reinbacher
developed a Monte Carlo implementation for quintic hypersurfaces and used
the volume-form error diagnostic $\sigma$ \cite{Douglas2008}.
Headrick and Wiseman solved a discretized Monge--Amp\`ere equation on K3
surfaces using lattice relaxation \cite{HeadrickWiseman2005}.
Headrick and Nassar developed energy-functional optimization within the
algebraic-metric framework \cite{HeadrickNassar2013}.
Braun, Brelidze, Douglas and Ovrut used numerical metrics to compute
eigenvalues and eigenfunctions of the scalar Laplacian
\cite{BraunSpectrum2008}. These methods approximate solutions of a
nonlinear equation while preserving K\"ahler geometry and global
consistency across coordinate patches.

Machine-learning approaches complement these classical constructions.
Ashmore, He and Ovrut combined machine learning with Donaldson's algorithm
\cite{Ashmore2020}. Ashmore, Calmon, He and Ovrut also learned K\"ahler
potentials from energy-optimized algebraic teachers \cite{AshmoreCalmon2023}.
Anderson et al. studied machine-learned metrics with moduli dependence
\cite{Anderson2021}. Jejjala, Mayorga Pe\~na and Mishra developed neural
network approximations to Calabi--Yau metrics \cite{Jejjala2022}, while
Douglas, Lakshminarasimhan and Qi investigated holomorphic-network
representations \cite{DouglasLakshminarasimhanQi2022}.

The \texttt{cymetric} framework supports neural metric constructions across
K\"ahler and complex-structure moduli
\cite{LarforsSizeShape2021,LarforsNumerical2022}.
CYJAX supplies JAX-based algebraic-metric calculations
\cite{GerdesKrippendorfCYJAX}, while \texttt{cymyc} provides tools for
numerical metrics, tensor fields and Yukawa couplings \cite{ButbaiaCymyc2025}.

Recent work has used learned potentials to obtain compact analytic
approximations. Mirjani\'c and Mishra used extrinsic symmetries and symbolic
regression to obtain compact
approximations to K\"ahler-potential corrections \cite{MirjanicMishra2025}.
Lee and Lukas fitted machine-learned potentials to a symmetry-reduced,
low-degree Donaldson ansatz for Dwork quintics and a family of bicubic
threefolds. Analytic fits to the coefficients made the complex-structure
dependence explicit \cite{LeeLukas2026}.

Constantin, Lukas and Nutricati used symbolic regression to fit
invariant-ansatz coefficients as functions of K\"ahler moduli
\cite{ConstantinLukasNutricati2026}. They studied bicubic and bi-degree
$(2,4)$ hypersurfaces at fixed complex structure. Constantin, Lee, Lukas
and Nutricati then combined K\"ahler- and complex-structure dependence
within an analytic potential ansatz \cite{ConstantinLeeLukasNutricati2026}.
Their demonstration varied K\"ahler moduli over a one-parameter
complex-structure family of bicubic threefolds.

Balanced fixed-point constructions and variationally optimized algebraic
metrics select different members of a finite-degree family.
Headrick and Nassar already restrict their polynomial search space to
symmetry-invariant combinations modulo the hypersurface equation
\cite[Sec.~6.1]{HeadrickNassar2013}. We use these established constructions
and build on
CYJAX's section evaluation and differential-geometric machinery
\cite{GerdesKrippendorfCYJAX}; the positive representative-block encoder
is the control parameterization used in this study. Symmetry-aware metric
construction itself is established prior work.

Algebraic coefficient matrices, potential corrections $\phi$ and metric
components are different learning targets. We regress the centered
volume-form error of an already computed metric. Predicting this error does
not reconstruct $\phi$ or test its conjectured ambient symmetries.

A scalar diagnostic can depend simply on the coordinates even when the
underlying metric is difficult to compute. An accurate fit to that
diagnostic may recover known geometry without revealing further metric
structure. Its residual can also contain repeatable features of the
numerical approximation. We ask which dependence is fixed by geometry
and which fitted patterns survive changes in representation or repeated
optimization.

We use an exact reference identity and finite Fermat symmetry to control
this comparison. Polynomial regressions fit the residuals of algebraic
metrics obtained by variational optimization. Replication and transfer
tests assess the stability of these fits under stochastic refinement and
changes in degree. We train no neural networks and do not reconstruct
the full metric from the scalar fits.

We study these questions on the Fermat quintic. For an algebraic metric
$g^{(k)}$ and the induced Fubini--Study reference, define
\begin{equation}
R^{(k)}=\log\frac{\det g^{(k)}}{\det g_{\mathrm{FS}}}.
\label{eq:ratio}
\end{equation}
The Ricci-flat volume-form relation fixes the exact reference to
\begin{equation}
R_\infty=C-\log p_4,\qquad
p_4=\sum_{i=0}^4|z_i|^8,\qquad \sum_i|z_i|^2=1.
\label{eq:fermat-reference}
\end{equation}
This is the Fermat specialization of the hypersurface identity of
Mirjani\'c and Mishra \cite[Theorem~3.4 and Appendix~A.1]{MirjanicMishra2025}.
We include a self-contained convention check and study the remaining field
\begin{equation}
E_k=R^{(k)}+\log p_4-c_k,
\label{eq:finite-residual}
\end{equation}
with its normalization $c_k$ determined on training points.
Here $R_\infty$ denotes the exact Ricci-flat reference, not an established
limit of the computed checkpoint sequence.

Symmetry is a second control. Fundamental-domain projections and
extrinsic-symmetry constructions organize numerical and symbolic
Calabi--Yau approximations \cite{Hendi2025,MirjanicMishra2025}.
GlobalCY compares global-invariant and symmetry-aware neural models with
local-input baselines on projective hypersurfaces \cite{RahmanGlobalCY2026}.
A fixed-modulus neural ablation also illustrates the sensitivity of
determinant diagnostics to geometric priors \cite{Eng2026Priors}.
Here we enforce the finite Fermat symmetry directly in the algebraic
coefficient matrix at every optimization step.
The experiment compares degrees $k=6,8,10,12$, repeats stochastic refinement
at fixed degree, tests whether polynomial residual fields transfer
across degrees, and evaluates one explicit phase-sensitive invariant
candidate at fixed degree.

Residual fields agree closely between the tested refinement seeds at fixed
$k$ within the related warm-start protocol, yet source-derived polynomial
models transfer poorly across degrees. Two further independent-start $k=10$
teachers attain comparable volume-form accuracy after a longer fixed budget.
Their fitted cubic modes agree much more closely than their complete
residual fields and retain predictive accuracy in cross-start transfer.
On a common line-grouped cloud, however, those modes agree poorly with the
principal warm-start teacher at the same degree. The pair's agreement
therefore does not extend to all tested constructions. Fitted functions
can remain similar even when the complete
teacher errors differ. These comparisons concern the retained checkpoints;
they identify neither a globally optimal algebraic residual nor an
asymptotic expansion as $k\to\infty$.

Calabi--Yau metrics enter compactification calculations. We examine the
reliability of learned patterns in their numerical approximations, without
computing a particle spectrum, effective theory or observable prediction.
The tests distinguish dependence fixed by geometry from patterns that
change with the representation or optimization procedure.

\section{Geometric Reference and Finite-Teacher Residual}

\subsection{Fermat geometry and normalization}

We work on the Fermat quintic
\[
X_0:\qquad \sum_{i=0}^{4}z_i^5=0
\]
in projective space. We use $Z$ for an arbitrary nonzero homogeneous
representative and $z=Z/\|Z\|$ for a unit-norm representative.
On a normalized representative,
$\sum_i |z_i|^2=1$, write $u_i=|z_i|^2$. The quantity
$p_4=\sum_i u_i^4$ is invariant under coordinate permutations and is strictly
positive on the normalized locus.

\begin{proposition}[analytic reference at the Fermat point]
\label{prop:fermat-reference}
For the normalized Fermat quintic, the Ricci-flat log-determinant ratio relative
to the Fubini--Study reference has the form in
Eq.~\eqref{eq:fermat-reference}.
\end{proposition}

See \cite[Theorem~3.4]{MirjanicMishra2025} for the general volume-ratio identity.
This Fermat specialization fixes the normalizations used below and is not
asserted as a new general determinant theorem.

\begin{proof}
Fix a projective chart $Z_a\neq 0$, write $w_j=Z_j/Z_a$ for $j\neq a$, and
choose one coordinate $w_d\neq 0$ to eliminate.  The remaining three affine
coordinates are denoted by $x_1,x_2,x_3$.  The defining equation becomes
\[
f(w)=1+\sum_{j\neq a}w_j^5=0.
\]
The Poincar\'e residue representative of the holomorphic three-form is,
up to a constant phase,
\[
\Omega=\frac{dx_1\wedge dx_2\wedge dx_3}{\partial f/\partial w_d}
=\frac{dx_1\wedge dx_2\wedge dx_3}{5w_d^4},
\]
following the standard residue construction for smooth projective
hypersurfaces \cite{Griffiths1969}.  Hence its coordinate density is
$|\Omega|^2=(25|w_d|^8)^{-1}$.

Let $S=1+\sum_{j\neq a}|w_j|^2$ and
$D=1+\sum_{j\neq a}|w_j|^8$. Reorder the affine coordinates so that
$w=(x_1,x_2,x_3,w_d)^{\mathsf T}$, and set
\[
v_r=-\frac{x_r^4}{w_d^4},\qquad
T=\begin{pmatrix}I_3\\v^{\mathsf T}\end{pmatrix},\qquad
M=T^{\mathsf T}\overline T=I_3+v\overline v^{\mathsf T}.
\]
The columns of $T$ are the holomorphic coordinate tangent vectors.
For the Fubini--Study potential $\log S$, the ambient matrix is
$G=S^{-1}I_4-S^{-2}\overline w\,w^{\mathsf T}$. Writing
$u=T^{\mathsf T}\overline w$, its pullback and the matrix determinant
lemma give
\begin{align}
T^{\mathsf T}G\overline T&=S^{-1}M-S^{-2}uu^\dagger,\nonumber\\
\det(T^{\mathsf T}G\overline T)
 &=S^{-4}\det(M)\bigl(S-u^\dagger M^{-1}u\bigr).
\label{eq:det-lemma-reference}
\end{align}
Since $\det M=1+\|v\|^2$ and
$M^{-1}=I_3-v\overline v^{\mathsf T}/(1+\|v\|^2)$, expansion yields
\[
\det(M)\bigl(S-u^\dagger M^{-1}u\bigr)
=1+\|v\|^2+\bigl|w_d-x^{\mathsf T}v\bigr|^2.
\]
The Fermat equation implies
$x^{\mathsf T}v=-\sum_r x_r^5/w_d^4=w_d+w_d^{-4}$. Hence
\[
1+\|v\|^2+\bigl|w_d-x^{\mathsf T}v\bigr|^2
=1+\frac{\sum_r|x_r|^8+1}{|w_d|^8}
=\frac{D}{|w_d|^8}.
\]
It follows that
\begin{equation}
\det(\iota^*g_{\mathrm{FS}})
=\frac{D}{|w_d|^8S^4}.
\label{eq:det-induced-reference}
\end{equation}
Consequently,
\[
\frac{|\Omega|^2}{\det(\iota^*g_{\mathrm{FS}})}
=\frac{S^4}{25D}
=\frac{\|Z\|^8}{25\sum_{i=0}^{4}|Z_i|^8}
=\frac{1}{25p_4},
\]
where the last equality uses the unit-norm representative.  The
complex Monge--Amp\`ere equation for the Ricci-flat metric fixes its volume
form to a positive constant multiple of $\Omega\wedge\overline{\Omega}$.
Therefore
\[
\frac{\det g_{\mathrm{RF}}}{\det g_{\mathrm{FS}}}
=\frac{C_0}{p_4}
\]
for a normalization constant $C_0>0$.  Taking logarithms proves
Eq.~\eqref{eq:fermat-reference}, with $C=\log C_0$.
\end{proof}

\begin{remark}[coordinate convention check]
The identity above concerns the volume form, not the full metric.  We evaluate
the residue density and induced Fubini--Study determinant independently in all
twenty regular choices of projective chart and eliminated coordinate on
thirty-two generic Fermat points.  The largest float64 relative discrepancy is
$1.3\times10^{-7}$, arising in poorly conditioned charts.  This checks the
power of $p_4$ and the determinant convention used here; it does not test the
convergence of any finite algebraic teacher.  A constant rescaling convention
for the Fubini--Study metric changes $C$, but cannot change the $-\log p_4$
dependence.
\end{remark}

\subsection{The residual is a volume-form error}

Let $\eta_k=\det(g^{(k)})/|\Omega|^2$ in compatible local coordinates.
The proof of Proposition~\ref{prop:fermat-reference} gives
$R^{(k)}+\log p_4=\log\eta_k-\log25$.
Consequently,
\begin{equation}
E_k=\log\eta_k-\langle\log\eta_k\rangle_{\mathrm{train},\mu}.
\label{eq:log-volume-error}
\end{equation}
Volume-form conventions change an additive constant, which cancels here.
Thus the quantity being compressed is the spatial variation of the
logarithmic volume-form error.

For $\eta_k/\langle\eta_k\rangle_\mu=1+\delta_k$ with uniformly small
$|\delta_k|$, Taylor expansion gives a centered linear contribution
$\delta_k-\langle\delta_k\rangle_{\mathrm{train},\mu}$ plus a centered
quadratic remainder. This interpretation does not follow from small
$\sigma$ alone: a small mean absolute error does not control the maximum.
An $L^1$ diagnostic and a logarithmic $L^2$ residual can also order two
teachers differently, especially under different sampling measures.

\section{Symmetry-Restricted Algebraic Metrics}
\label{sec:invariant}

\subsection{Section representation}

Let $s(z)$ be the column of degree-$k$ monomial sections reduced modulo
$\sum_i z_i^5=0$. For the degrees used here,
\[
N_k=\binom{k+4}{4}-\binom{k-1}{4}.
\]
We use the convention
\begin{equation}
K_H=\frac1k\log\bigl(s(z)^{\mathsf T}H\overline{s(z)}\bigr),\qquad
g_H=\partial\bar\partial K_H,\qquad H=H^\dagger>0.
\label{eq:algebraic}
\end{equation}
Local potentials are obtained by trivializing the section bundle; their
metrics agree on overlaps. An overall multiplicative convention for $g_H$
changes only the normalization of the determinant reference.
With this fixed $1/k$ convention, the complete section embedding represents
the same polarization class at every degree.

The finite Fermat symmetry is
$G=((\mathbb Z/5\mathbb Z)^5/\mathbb Z/5\mathbb Z)\rtimes S_5$.
The diagonal quotient acts projectively on sections and by a well-defined
congruence on $H$. At fixed total degree, monomials have the same diagonal
character precisely when their exponent vectors agree modulo five.
Diagonal invariance therefore makes $H$ block diagonal in those sectors.
For $s(gz)=A_gs(z)$, permutation invariance requires
\begin{equation}
A_g^{\mathsf T}H\overline{A_g}=H.
\label{eq:congruence}
\end{equation}
The reduced basis eliminates powers containing $z_0^5$.
A permuted section must be reduced in the quotient ring, so $A_g$
is generally not a permutation matrix.

\subsection{Positive invariant construction}

For each $S_5$ orbit $\mathcal O$ of character sectors, choose a
representative sector of dimension $m_{\mathcal O}$ and a positive block
$B_{\mathcal O}=L_{\mathcal O}L_{\mathcal O}^\dagger$.
Let $J_{\mathcal O}$ inject that sector into the full section space.
Define
\begin{equation}
H(B)=\frac1{120}\sum_{\mathcal O}\sum_{g\in S_5}
 A_g^{\mathsf T}J_{\mathcal O}B_{\mathcal O}J_{\mathcal O}^{\mathsf T}\overline{A_g}.
\label{eq:invariant-h}
\end{equation}
The implementation computes this average in character blocks, retaining
all transports, including stabilizer multiplicities.

\begin{proposition}[positivity and finite symmetry]
\label{prop:invariant-h}
For positive representative blocks, Eq.~\eqref{eq:invariant-h} is positive
definite and invariant under $G$. Arbitrary positive representative blocks
cover the positive invariant cone, although the map need not be one-to-one.
\end{proposition}

\begin{proof}
Each summand is positive semidefinite. Every character sector receives
positive-definite congruences of its representative block because its
permutation transports are invertible. The complete block-diagonal matrix
is therefore positive definite. Each summand is diagonal-character
invariant, and a further permutation merely reorders the group average.
For surjectivity, take a positive invariant $H_*$.
On an orbit of size $|\mathcal O|$, choose
$B_{\mathcal O}=|\mathcal O|(H_*)_{\mathcal O\mathcal O}$ on the
representative sector. There are $120/|\mathcal O|$ transports to each
target sector. Invariance makes their averaged contribution exactly the
target block of $H_*$. Summing over orbits recovers $H_*$.
\end{proof}

Complex triangular factors with positive real diagonals use
$\sum_{\mathcal O}m_{\mathcal O}^2$ real optimization coordinates
(Table~\ref{tab:dimensions}). These counts are not the dimension of the
invariant cone: stabilizer averaging and the irrelevance of an overall
scale can introduce redundancy. The numerical encoder uses softplus
diagonals with a small positive floor; the surjectivity statement concerns
the unrestricted positive-block construction.
An overall positive scale of $H$ leaves the metric unchanged. The retained
optimization used these redundant coordinates without an explicit
scale-gauge fixing or minimal stabilizer-coordinate reduction.
The exact symmetry statement concerns the specified holomorphic group $G$.
No separate anti-holomorphic conjugation constraint was imposed or audited
for these teachers.

\begin{table}[H]
\centering
\caption{Section dimensions and real optimization-coordinate counts.
The largest block is a character-sector block before stabilizer averaging.}
\label{tab:dimensions}
\begin{tabular}{rrrrr}
\hline
$k$ & $N_k$ & character orbits & largest block & coordinates\\
\hline
6 & 205 & 9 & 4 & 24\\
8 & 460 & 14 & 4 & 59\\
10 & 875 & 19 & 10 & 208\\
12 & 1490 & 22 & 10 & 355\\
\hline
\end{tabular}
\end{table}

\section{Computational Protocol}

Random-stream identifiers and implementation records are given in
Appendix~\ref{app:reproducibility}.

\subsection{Optimization and diagnostic}

The retained teachers were refined inside Eq.~\eqref{eq:invariant-h}
for 30,000 Adam updates in float64 arithmetic. Each update accumulated
four microbatch gradients from 128 newly sampled points each.
A cosine schedule decreased the learning rate from $3\times10^{-4}$
to $10^{-5}$ over the refinement.
Initialization used symmetry-projected checkpoints at the same
degree. The 30,000 updates describe this refinement, not the entire
training cost.

For a batch of $B=128$ points, let $w_i$ be the positive importance
weights defined in Eq.~\eqref{eq:importance-weight} and
$\eta_i=\det g_H/|\Omega|^2$ at point $i$.
The implemented microbatch loss was
\[
\widehat{\mathcal L}
=\frac1B\sum_{i=1}^B w_i
 \left(\frac{\eta_i}{\bar\eta_w}-1\right)^2,\qquad
\bar\eta_w=\frac{\sum_iw_i\eta_i}{\sum_iw_i}.
\]
The outer average uses the sampled weights as supplied, without dividing
by their batch mean; the ratio defining $\bar\eta_w$ is self-normalized.
We held $\bar\eta_w$ fixed during differentiation and averaged the gradients
of four separately normalized microbatch losses. Normalization was not
computed on a single 512-point batch. This stop-gradient update generally
differs from the total derivative of the displayed finite-microbatch loss
because $\bar\eta_w$ depends on $H$. Fixed total metric volume in a
K\"ahler class does not make the finite-batch update an unbiased exact
variational gradient. The results concern this stochastic rule, with no
claim of an implementation-independent solver for the classical energy
functional. The training also differs from solving Donaldson's balanced
fixed-point equation.

With $d\mu_\Omega$ the positive Calabi--Yau volume measure, we report
\begin{equation}
\sigma(H)=\frac1{\operatorname{Vol}_\Omega(X)}
\int_X\left|1-\frac{\eta_H}{\langle\eta_H\rangle_{\mu_\Omega}}\right|
\,d\mu_\Omega.
\label{eq:sigma}
\end{equation}
Unless stated otherwise, checkpoint means and sample standard deviations used
five fresh clouds of 10,000 points per checkpoint. Those standard deviations measure
evaluation-cloud variation conditional on a checkpoint.

We repeated the principal refinements at $k=8,10,12$ with different
stochastic streams. These runs test repeatability within the related
warm-start protocol; they do not sample arbitrary initializations. The
$k=12$ pair shares a projected starting matrix. A separate $k=10$ control
started from a Gaussian perturbation of the invariant identity at scale
$0.05$, retaining the invariant parameterization throughout.
Final reported $S_5$ matrix residuals were below $10^{-15}$;
diagonal-character invariance is built into the block structure.

\subsection{Common clouds and regressions}

Random complex-line intersections with the quintic supplied normalized
points. All compared teachers were evaluated at the same points, in
compatible local coordinates. Both metric determinants transform by
$|\det J|^2$, making their ratio coordinate independent.
The four principal teachers and their three warm-start replicas share
50,000 points on 10,000 complete five-root lines. An 80:20 line split
supplies 40,000 training and 10,000 held-out points, on 8,000 and 2,000
lines respectively. The principal mode-space analysis uses only the four
principal teachers. Each replication analysis uses its same-degree pair;
a same-degree replica never enters a cross-degree source span.

For each measure $\mu$, $c_k$ was the weighted training mean of
$R^{(k)}+\log p_4$. The regression included all monomials of total
feature degree at most three in $(p_2,p_3,p_4,p_5)$, where
$p_j=\sum_i|z_i|^{2j}$. This gives 35 coefficients including an intercept.
Cubic degree refers to degree in these four features, not degree in
homogeneous coordinates.
The elementary symmetric generators do not supply independent geometric
information. For example, $e_3=(1-3p_2+2p_3)/6$ on the normalized locus.
Changing generators can nevertheless change a truncated library and its
numerical conditioning.
Before forming monomials, each of the four power sums was centered by its
unweighted training mean and divided by its unweighted training standard
deviation (population convention, divisor equal to the training count).
All monomials through the specified degree were then formed from these
standardized generators, including the constant column; the monomial
columns were not separately rescaled. The appended $J_7$ column used its
own unweighted training mean and standard deviation. These conventions
were unchanged for volume-weighted least squares, whose design rows and
targets were multiplied by the square roots of the training weights.
Offsets, feature scaling, coefficients and transfer amplitudes used
training points only.

\subsection{Phase-sensitive candidate and modulus-only controls}
\label{sec:j7-method}

To probe phase dependence omitted by the modulus-only cubic library,
we used the normalized Fermat-invariant candidate
\begin{equation}
J_7(z)=\sum_{i\ne j}z_i^5\overline{z_j}^5|z_i|^2|z_j|^2,
\qquad \sum_i|z_i|^2=1.
\label{eq:j7}
\end{equation}
It is real because the $(i,j)$ and $(j,i)$ summands are complex conjugates.
Equivalently,
\[
J_7(z)=\left|\sum_i z_i^5|z_i|^2\right|^2-\sum_i|z_i|^{14}.
\]
It is invariant under the diagonal fifth-root action because $z_i^5$ is
unchanged, and it is invariant under $S_5$ because the ordered-pair sum is
permuted. Its phase dependence requires an additional argument.
Take unnormalized fifth powers
$(e^{ia},e^{-ia},2e^{ib},2e^{-ib},-1)$, with
$\cos b=(1-2\cos a)/4$ and $0<a<\pi$.
Their sum is zero and the coordinate magnitudes do not depend on $a$.
Writing $S=3+2\cdot 2^{2/5}$ for the squared norm before normalization,
direct substitution into Eq.~\eqref{eq:j7} gives
\[
J_7(a)=\frac{(1-2^{2/5})^2(2\cos a-1)^2-(3+2^{19/5})}{S^7}.
\]
This is nonconstant, for example between $a=\pi/3$ and $a=\pi/2$.
Thus $J_7$ is not a function of the moduli alone and cannot belong to the
cubic power-sum library. This does not make it a first or complete
phase-sensitive invariant. Nonmembership in a modulus-only library does
not itself imply predictive value for a particular residual.

The feature was selected exploratorily. We compare the 35-column cubic
and 70-column quartic modulus-only libraries with their respective
one-column, training-standardized $J_7$ extensions on the common clouds.
For same-degree transfer, the source fitted mode was frozen; only one scalar
amplitude was fitted on the target training residual before held-out scoring.
The function libraries are nested, but rank truncation need not preserve
exact nesting of the retained numerical subspaces. Reported increments
compare the recorded fitting rules, not ideal full-rank projections.
Improvement over a cubic library alone does not isolate
irreducible conditional phase information: $J_7$ can also correlate with
modulus dependence missed by that library. The quartic controls assess
this dependence on the chosen modulus-only feature space.
To separate candidate augmentation from changes in numerical rank, a
fixed-space control holds the retained weighted baseline basis $Q$ and
coefficient map $M$ fixed. The candidate is centered and scaled using
weighted training moments, projected off that baseline, and reorthogonalized
once. If its remaining weighted training norm exceeds $10^{-8}$ of its
original norm, only that additional direction is fitted to the baseline
residual. Otherwise the baseline prediction is returned unchanged.
This preserves the baseline projector and prediction rather than
recomputing an SVD of the augmented design. The same line splits and
1,000 paired test-line bootstrap draws are used at each of the three cutoffs.

\subsection{Sampling measures and uncertainty}

We evaluated the same points under the native intersection measure and
importance weights for $\mu_\Omega$. The resulting statistics are not
independent geometric samples. Let $d\mu_{\rm int}$ denote the measure
induced by complex-line intersection sampling. The importance
ratio
\begin{equation}
w(z)=\frac{d\mu_\Omega}{d\mu_{\rm int}}(z)
=\frac{1}{\det g_{\rm FS}^{\rm ind}(z)\,|\partial_{w_d}f(z)|^2},
\label{eq:importance-weight}
\end{equation}
is defined up to a global constant. Both denominator factors are evaluated
in the same affine chart $w_j=z_j/z_a$ with $z_a\ne0$, not by substituting
unit-normalized homogeneous components into an affine derivative.
Here $w_d$ is the locally eliminated coordinate and
$g_{\rm FS}^{\rm ind}$ is the induced Fubini--Study matrix in the three
remaining coordinates. Proposition~\ref{prop:fermat-reference} gives the
chart-independent expression $w(z)\propto [25p_4(z)]^{-1}$ on the
unit-normalized Fermat quintic. Global normalization constants cancel from the
statistics below.
For a finite split $I$ and positive weights $w_i$, write
\begin{equation}
\langle a\rangle_w=\frac{\sum_{i\in I}w_i a_i}{\sum_{i\in I}w_i},
\qquad
\|a\|_w^2=\frac{\sum_{i\in I}w_i a_i^2}{\sum_{i\in I}w_i}.
\label{eq:weighted-mean-norm}
\end{equation}
The weighted held-out statistics are
\begin{align}
R_w^2(y,\widehat y)&=1-
\frac{\sum_{i\in I}w_i(y_i-\widehat y_i)^2}
     {\sum_{i\in I}w_i(y_i-\langle y\rangle_w)^2},
\label{eq:weighted-r2}\\
\operatorname{corr}_w(a,b)&=
\frac{\sum_{i\in I}w_i(a_i-\langle a\rangle_w)(b_i-\langle b\rangle_w)}
{\sqrt{\sum_{i\in I}w_i(a_i-\langle a\rangle_w)^2}
 \sqrt{\sum_{i\in I}w_i(b_i-\langle b\rangle_w)^2}}.
\label{eq:weighted-correlation}
\end{align}
Multiplying all weights by the same positive constant leaves these
quantities unchanged. The unweighted intersection results set $w_i=1$.
The seven-teacher and three-teacher regression evaluations use disjoint
training and test lines and 1,000 paired test-line bootstrap replicates.
Each sampled line contributes all five roots, retaining teacher pairing
and importance weights. The 95\% percentile intervals condition on the
fixed teachers, feature libraries and fitted models. They exclude
optimization and model-selection uncertainty and are not simultaneous
across all comparisons. Feature selection and the quartic mode-space
sensitivity analysis are exploratory.

\subsection{Shared-mode tests}
\label{sec:mode-method}

Scalar transfer fitted a cubic polynomial to one teacher, removed its
training mean and froze its shape. Only an amplitude was fitted to the
target teacher's training residual before held-out evaluation.

For joint modes, let $X$ be the polynomial design, $W$ the diagonal training
weights and $Y$ the matrix of centered teacher residuals.
For a full-column-rank design, a thin factorization $W^{1/2}X=QT$
gives an orthonormal basis $Q$ and a coefficient map $M=T^{-1}$.
For a numerically rank-deficient design, write the retained SVD as
$U_q\Sigma_q V_q^{\mathsf T}$ and instead set
$Q=U_q$ and $M=V_q\Sigma_q^{-1}$, so that $W^{1/2}XM=Q$.
In either case the functional coordinates are $C=Q^{\mathsf T}W^{1/2}Y$.
Rank-$r$ truncation $C_r$ gives coefficients $MC_r$ and predictions
$XMC_r$ on new points. The design basis and map $M$ remain fixed when
rescaling teacher columns. The SVD uses the training function
norm, avoiding dependence on the coefficient norm of the nonorthogonal
monomial basis. Full rank recovers the individual polynomial fits.
For full-rank designs the implementation uses a linear solve, not an
explicit matrix inverse. The analysis uses a rank-truncated SVD
and its coefficient map if the design is numerically rank deficient,
retaining only singular values above the recorded cutoff. It records
the singular spectrum, numerical rank, relative rank cutoff and condition
number of each weighted design. Least-squares cutoff sensitivity can be
checked without changing the teachers. Cubic and quartic $J_7$ cutoff
checks are retained with both line-grouped evaluations.

Table~\ref{tab:conditioning} summarizes the weighted baseline designs.
The default relative cutoff is $40000\epsilon_{64}$, approximately
$8.88\times10^{-12}$. At cutoffs $10^{-10}$ and $10^{-12}$, the seven-teacher
cubic designs retain all 35 directions and the quartic designs retain
62--69 directions. Numerical rank is not an exact algebraic rank.

\begin{table}[tb]
\centering
\small
\caption{Conditioning of the modulus-only regression designs, aggregated
over teachers and measures within each cloud. Ranks use the default cutoff;
$\kappa$ is the ratio of largest to smallest singular value before truncation.
Full singular spectra and cutoff-specific ranks are retained in the analysis records.}
\label{tab:conditioning}
\begin{tabular}{llrr}
\hline
cloud & library & rank / columns & $\kappa$ range\\
\hline
seven-teacher & cubic & 35 / 35 & $[2.30,3.65]\times10^9$\\
seven-teacher & quartic & 66--67 / 70 & $[6.18,10.04]\times10^{12}$\\
three-teacher & cubic & 35 / 35 & $[2.37,3.69]\times10^9$\\
three-teacher & quartic & 66--67 / 70 & $[6.57,10.64]\times10^{12}$\\
\hline
\end{tabular}
\end{table}

We also divided each residual column by its training RMS before fitting,
then restored its units for prediction.
Singular energy fractions concern the training polynomial fits, not
held-out total residual variance. Joint RMSE remains in original residual
units, including for the normalized fit.

The source subspace uses only the other three degrees.
Target training points determine its coordinates in
that frozen subspace; target test points determine its score.
This tests reuse of a functional basis, not prediction at a new degree
without target data.
A post-hoc quartic mode-space analysis uses the same principal cloud,
split and numerical-rank rule. It assesses library sensitivity, rather
than selecting the best library by held-out performance.

\begin{proposition}[rescaling and full source-span transfer]
\label{prop:span}
Projection onto the full column span of the source coordinate matrix
is unchanged by nonzero rescaling of its individual columns.
Rescaling the target and restoring its scale after projection also leaves
the prediction unchanged.
\end{proposition}

\begin{proof}
For an invertible diagonal matrix $D$,
let $C_{-j}$ denote the source coordinate columns, excluding target $j$,
and let $c=C_{:j}$ denote that target's coordinate vector. Then
$\operatorname{col}(C_{-j}D)=\operatorname{col}(C_{-j})$.
The orthogonal projector $P$ onto this space is therefore unchanged,
and linearity gives $aP(c/a)=Pc$ for a nonzero target scale $a$.
Mapping through the same fixed $M$ and polynomial design preserves equality,
including when $M$ is the retained-SVD coefficient map.
\end{proof}

When the three source fields are independent, raw and normalized rank-three
transfer must agree. Ranks one and two can differ because rescaling changes
the directions prioritized by the SVD. If sources are rank deficient,
only their nonzero singular directions define the source span.

\subsection{Continuation control}
\label{sec:continuation-method}

To distinguish late-stage optimization changes from degree changes, we
restarted Adam once from the retained invariant $H$ matrices at $k=10$ and
$k=12$, using the same invariant parameterization throughout. The
continuation used 30,000 additional updates, four microbatches
of 128 points per update, and a cosine learning-rate schedule from
$10^{-4}$ to $10^{-5}$. The optimizer state was reset only at the retained
baseline; snapshots after 10,000, 20,000 and 30,000 updates were taken from
one uninterrupted refinement stream.

For a stage $s$ and the other degree $\ell$, the audit compared the held-out
residual differences
\begin{equation}
q^{(0)}_{k,s;\ell}
=\frac{\|E_{k,s}-E_{k,0}\|_\mu}
        {\|E_{k,s}-E_{\ell,s}\|_\mu},\qquad
q^{(\mathrm{step})}_{k,s;\ell}
=\frac{\|E_{k,s}-E_{k,s-10{,}000}\|_\mu}
        {\|E_{k,s}-E_{\ell,s}\|_\mu}.
\label{eq:continuation-ratios}
\end{equation}
The first ratio measures cumulative movement from the retained baseline;
the second measures the most recent continuation step. Both numerator and
denominator use the same common-cloud split and measure. A 10,000-point
cloud and an independent 50,000-point cloud provide descriptive comparisons.
These controls use individual-point splits, not line-disjoint holdout;
no uncertainty intervals are assigned. The ratios describe movement under
the annealed schedule, not bounds on optimization error.

A 30,000-update initialization control is described in
Appendix~\ref{app:initialization-control}.

\subsection{Matched-quality independent starts}
\label{sec:matched-independent-method}

A separate fixed-budget experiment trained two $k=10$ teachers from
independently seeded Gaussian perturbations of the invariant identity,
at scale $0.05$, without loading a learned checkpoint. Both initialization
and stochastic training varied, so their effects are not isolated.
Both runs used the same invariant
parameterization, objective, and four 128-point microbatches per update.
Each completed 90,000 updates on one cosine schedule from $3\times10^{-4}$
to $10^{-5}$. Final matrices were selected by the fixed update budget,
not by residual agreement.

An operational accuracy check used ten fresh paired 10,000-point clouds
to estimate each final mean sigma. Both means had to be at most $0.05\%$,
with the larger divided by the smaller at most 1.20. Passing this check
permits comparison at similar volume-form accuracy; it does not certify
convergence or equal residual tails.

Both endpoints were evaluated with the principal teacher on a separate
50,000-point line-grouped cloud, as specified below.
An operational agreement criterion required, under both measures, a
full-residual correlation interval lower bound
of at least 0.95. Both cross-start transfer scores also had to be positive
and no more than 0.05 below the target's own cubic-mode score, using
held-out $R^2$ point estimates for this transfer-loss threshold.
This is an operational shape-agreement and predictive-transfer criterion,
not a population equivalence test or a bound on relative field difference.
Correlation does not constrain amplitude, and a matched sigma does not
match logarithmic residual norms. Complete residuals and fitted modes are
assessed separately.

\subsection{Line-grouped linkage at \texorpdfstring{$k=10$}{k=10}}
\label{sec:line-review-method}

The principal warm-start teacher and the two independent endpoints were
evaluated on a common cloud of 50,000 points.
All five roots of each of 10,000 random complex lines were retained.
The 80:20 split separated 8,000 training lines from 2,000 test lines,
with no parent line shared between splits. Any invalid root would reject
its entire line; no points were rejected in this evaluation.
Offsets, scaling, regression coefficients and transfer amplitudes used
training lines only. The 1,000 paired bootstrap draws resampled complete
test lines, retaining all teachers and importance weights together.

The same split compared the cubic library with its one-column $J_7$
extension, and the 70-column quartic modulus-only library with its own
$J_7$ extension. These are nested comparisons within each library.
Weighted numerical ranks and singular spectra were recorded. Additional
local checks used relative least-squares cutoffs $10^{-10}$ and $10^{-12}$
for both libraries. These checks assess sensitivity, not selection of the
best held-out cutoff. Evaluation did not retrain the teachers or isolate
initialization from training budget.

\section{Results}

\subsection{Reference accuracy and residual size}

The invariant teachers had small volume-form errors
(Table~\ref{tab:reference}), but their accuracy was non-monotone in $k$.
After the same refinement budget, the $k=12$ diagnostic exceeded those
of $k=8,10$. The analytic reference explained almost all variation of
$R^{(k)}$; its remaining residual nevertheless had resolvable structure.
Individual cubic fits explained 0.711--0.958 of held-out residual
variance under intersection sampling and 0.624--0.922 under volume
weighting.

\begin{table}[tb]
\centering
\caption{Principal invariant sequence on one 50,000-point common cloud,
with 10,000 held-out points on 2,000 disjoint test lines.
Residual RMSE equals the error of the
offset-fitted exact reference. The final two columns give held-out
residual-polynomial $R^2$. All teachers belong to the principal sequence.
$\sigma$ is a percentage from separate final evaluations.}
\label{tab:reference}
\small
\begin{tabular}{crrrrr}
\hline
$k$ & $\sigma$ (\%) & RMSE$_{\rm int}$ & RMSE$_\Omega$
 & $R^2_{\rm poly,int}$ & $R^2_{\rm poly,\Omega}$\\
\hline
6 & 0.1411 & 0.002482 & 0.002021 & 0.958455 & 0.921715\\
8 & 0.0490 & 0.000627 & 0.000632 & 0.893214 & 0.865281\\
10 & 0.0436 & 0.000650 & 0.000567 & 0.729694 & 0.653304\\
12 & 0.0641 & 0.000903 & 0.000823 & 0.710787 & 0.624140\\
\hline
\end{tabular}
\end{table}

\subsection{Movement under continuation}
\label{sec:continuation-results}

The continuation control produced a consistent separation between recent
within-degree movement and the contemporaneous cross-degree difference
(Table~\ref{tab:continuation}). On the independent 50,000-point cloud, the
final 10,000-update change was 0.019--0.021 of the cross-degree difference
for $k=10$ and 0.033 for $k=12$, under the two measures. The corresponding
values on the independent 10,000-point cloud were 0.020--0.022 and
0.035--0.036.

The cumulative change from the baseline checkpoint was larger: 0.167--0.187
for $k=10$ and 0.227--0.246 for $k=12$ on the 50,000-point cloud. The
continuation therefore made a measurable adjustment before entering a regime
of smaller successive changes under the annealed schedule. Decreasing the
learning rate can itself reduce movement. The comparison therefore does
not certify convergence or bound the distance to an optimum.

\begin{table}[tb]
\centering
\caption{Continuation ratios at the 30,000-update snapshot. $q^{(0)}$
measures cumulative change from the retained baseline and
$q^{(\mathrm{step})}$ measures the final 10,000-update change, each divided
by the matched cross-degree residual difference. The 10k and 50k columns
refer to independently sampled clouds. These are descriptive point-split
comparisons without line-cluster uncertainty estimates.}
\label{tab:continuation}
\small
\begin{tabular}{llrrrr}
\hline
$k$ & measure & $q^{(0)}_{10k}$ & $q^{(\mathrm{step})}_{10k}$
 & $q^{(0)}_{50k}$ & $q^{(\mathrm{step})}_{50k}$\\
\hline
10 & intersection & 0.182 & 0.020 & 0.167 & 0.019\\
10 & volume weighted & 0.191 & 0.022 & 0.187 & 0.021\\
12 & intersection & 0.240 & 0.036 & 0.227 & 0.033\\
12 & volume weighted & 0.249 & 0.035 & 0.246 & 0.033\\
\hline
\end{tabular}
\end{table}

Figure~\ref{fig:continuation} shows decreasing recent-step changes alongside
the cumulative movement from the baseline on the 10,000-point cloud.

\begin{figure}[tb]
\centering
\includegraphics[width=0.88\linewidth]{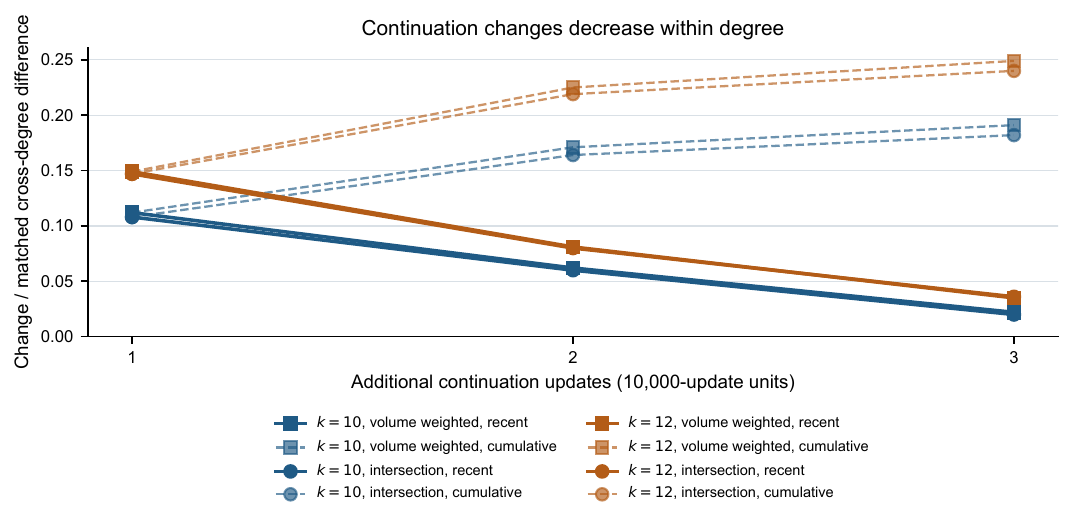}
\caption{Continuation trajectory on the 10,000-point cloud.
Solid curves show the most recent 10,000-update change divided by the matched
cross-degree difference; dashed curves show cumulative movement from the
retained baseline. Circles denote intersection sampling and squares denote
volume weighting. Each point is a paired held-out comparison, not an error
bar or a convergence bound.}
\label{fig:continuation}
\end{figure}

\subsection{Replication at fixed degree}
\label{sec:replication}

Changing the stochastic refinement seed within the related warm-start
protocol produced very similar residual fields at each replicated degree.
Same-degree correlations exceeded
0.9997 for $k=8,10,12$ under both measures
(Table~\ref{tab:replication}). This compares errors point by point,
beyond agreement of the scalar $\sigma$ values.

\begin{table}[tb]
\centering
\caption{Held-out residual correlations between paired warm-start
refinements on the shared 50,000-point cloud. Brackets give conditional
95\% paired line-bootstrap intervals on 2,000 held-out lines.}
\label{tab:replication}
\small
\begin{tabular}{crr}
\hline
$k$ & correlation$_{\rm int}$ & correlation$_\Omega$\\
\hline
8 & 0.999807 [0.999797, 0.999817] & 0.999860 [0.999852, 0.999868]\\
10 & 0.999781 [0.999765, 0.999798] & 0.999832 [0.999824, 0.999839]\\
12 & 0.999878 [0.999869, 0.999887] & 0.999916 [0.999912, 0.999920]\\
\hline
\end{tabular}
\end{table}

At $k=12$, the between-run residual difference had RMSE approximately
$1.4\times10^{-5}$ under intersection sampling and
$1.1\times10^{-5}$ under volume weighting, about 1--2\% of the residual
norms. Transferring the fitted cubic mode between these refinements preserved
the within-teacher scores: approximately 0.71 for intersection sampling
and 0.62 for volume weighting. These are scores on the same
cloud and split as Table~\ref{tab:reference}, not distribution-independent
constants.

\subsection{A phase-sensitive invariant contributes at fixed degree}
\label{sec:j7-results}

The one-column $J_7$ extension increased held-out scores beyond the cubic
baseline for all seven warm-start teachers under both measures. All 14
conditional 95\% paired line-bootstrap intervals excluded zero.
Table~\ref{tab:j7} reports the replicated $k=10,12$ comparisons.
The paired improvements differed by at most $6.8\times10^{-4}$, with
matching coefficient signs within each degree--measure pair.
For the principal $k=6$ teacher, $R^2$ increased from 0.9585 to 0.9912
under intersection sampling and from 0.9217 to 0.9845 under volume weighting.

The improvement also survived frozen transfer between the paired warm-start
refinements. A mode fitted to one teacher, including its $J_7$ component, improved
held-out $R^2$ on the other teacher after only a target-training scalar amplitude
was fitted. Thus the result is not merely a separate refit for each teacher
within this protocol. The independent-start comparison tests the boundary
of that statement. These results do not establish phase dependence of the
Ricci-flat metric, an invariant-ring classification, or transfer of this
component across degrees.

\begin{table}[tb]
\centering
\caption{Common-cloud $J_7$ augmentation for paired warm-started
teachers, with 10,000 held-out points. ``Own fit'' gives the paired
held-out improvements from adding the one standardized $J_7$ column to the
cubic power-sum baseline. ``Frozen transfer'' gives the improvement when the
source fitted mode is transferred in each direction and only a target-training
scalar amplitude is fitted. All eight within-teacher conditional 95\%
paired line-bootstrap intervals excluded zero. In each own-fit pair the
principal teacher precedes its replica; transfer pairs are ordered
principal $\to$ replica, then replica $\to$ principal.}
\label{tab:j7}
\small
\begin{tabular}{llrr}
\hline
measure & $k$ & own-fit $\Delta R^2$ & frozen-transfer $\Delta R^2$\\
\hline
intersection & 10 & 0.008710, 0.008798 & 0.008779, 0.008728\\
intersection & 12 & 0.006854, 0.007281 & 0.007292, 0.006830\\
volume weighted & 10 & 0.005142, 0.005192 & 0.005181, 0.005154\\
volume weighted & 12 & 0.010844, 0.011513 & 0.011488, 0.010848\\
\hline
\end{tabular}
\end{table}

The complete cubic own-fit results, including $k=6,8$ and all fourteen
conditional intervals, are given in Appendix~\ref{app:complete-j7}.

The larger modulus-only control changes the interpretation at $k=12$
(Table~\ref{tab:degree-quartic-j7}). Its quartic fits explain approximately
97--98\% of residual variance. Adding $J_7$ then gives gains of order
$10^{-5}$ or smaller, with all four conditional intervals containing zero
across the two replicas and measures. In contrast, quartic-baseline gains
remain positive at $k=6,8,10$. A gain beyond the cubic library therefore
does not establish irreducible conditional phase information. Nor does
the unresolved quartic increment at $k=12$ establish phase independence.

\begin{table}[tb]
\centering
\small
\caption{Quartic modulus-only control on the seven-teacher cloud.
Principal teachers are shown; both $k=12$ replicas have intervals containing
zero under both measures. Intervals are conditional 95\% paired test-line
bootstrap intervals for the one-column $J_7$ increment, not simultaneous bounds.}
\label{tab:degree-quartic-j7}
\begin{tabular}{lrrr}
\hline
measure & $k$ & quartic $R^2$ & $J_7$ increment [interval]\\
\hline
intersection & 6 & 0.965829 & 0.030575 [0.026138, 0.035219]\\
intersection & 8 & 0.931000 & 0.001108 [0.000680, 0.001537]\\
intersection & 10 & 0.893332 & 0.037976 [0.034151, 0.042047]\\
intersection & 12 & 0.978915 & $-0.000003\;[-0.000009, 0.000003]$\\
volume weighted & 6 & 0.933604 & 0.059540 [0.055291, 0.064170]\\
volume weighted & 8 & 0.903110 & 0.003861 [0.002872, 0.004807]\\
volume weighted & 10 & 0.868004 & 0.039393 [0.035950, 0.043203]\\
volume weighted & 12 & 0.974154 & $0.000003\;[-0.000044, 0.000050]$\\
\hline
\end{tabular}
\end{table}

\subsection{Matched-quality starts share a fitted component}
\label{sec:matched-independent-results}

The longer fixed-budget pair, denoted $A$ and $B$, passed the accuracy check.
The paired-cloud mean sigma was $0.035895\%$ for $A$ and $0.040421\%$ for $B$.
The corresponding sample standard deviations were $0.000432\%$ and
$0.000437\%$. The ratio of mean sigmas was 1.126, below the operational 1.20 limit.
Both final matrices retained $S_5$ invariance to relative residual below
$5\times10^{-16}$. These are separate starts from the principal warm-start
sequence, not further refinements of its checkpoints.

The complete residuals were positively correlated, but did not meet the
operational full-field agreement threshold
(Table~\ref{tab:matched-independent-correlation}). The cubic fitted modes
were more closely aligned. This is a comparison of functions on held-out
points, not an inference from coefficient similarity in a nonorthogonal
polynomial basis.

\begin{table}[H]
\centering
\caption{Matched-quality independent $k=10$ starts at 90,000 updates.
Complete-residual intervals are 95\% paired line-bootstrap intervals
on 2,000 held-out lines (10,000 points), conditional on the two teachers.
The scores use the three-teacher cloud of Section~\ref{sec:line-review-method}.
The fitted-mode statistic compares the prescribed cubic surrogate functions.}
\label{tab:matched-independent-correlation}
\small
\medskip
\begin{tabular}{lrr}
\hline
measure & residual correlation [95\% interval] & fitted-mode correlation\\
\hline
intersection & 0.8032 [0.7902, 0.8163] & 0.9831\\
volume weighted & 0.7677 [0.7567, 0.7788] & 0.9786\\
\hline
\end{tabular}
\end{table}

Cross-start frozen cubic modes retained useful held-out predictive accuracy
(Table~\ref{tab:matched-independent-transfer}). Each score was positive,
and its loss against the target's own cubic fit was between 0.024 and
0.029 in $R^2$. The source shape was frozen, but one amplitude used target
training values; this is not zero-shot prediction without target data.
The result supports cross-start reuse of a component inside the chosen
cubic library, rather than agreement of the entire residual.

\begin{table}[H]
\centering
\caption{Frozen cubic-mode transfer for the matched-quality pair $A,B$.
\emph{Own} is the target's own cubic-mode score; \emph{transfer} uses the other
start's frozen mode with one train-only amplitude calibration.
The loss is own minus transfer. All entries are held-out $R^2$.}
\label{tab:matched-independent-transfer}
\small
\medskip
\begin{tabular}{llrrr}
\hline
measure & source $\to$ target & own & transfer & loss\\
\hline
intersection & $A\to B$ & 0.6619 & 0.6359 & 0.0260\\
intersection & $B\to A$ & 0.7799 & 0.7558 & 0.0240\\
volume weighted & $A\to B$ & 0.5556 & 0.5274 & 0.0282\\
volume weighted & $B\to A$ & 0.6725 & 0.6471 & 0.0253\\
\hline
\end{tabular}
\end{table}

The pair passed the transfer criterion but failed the full-field agreement
criterion: both correlation intervals lie below 0.95. Agreement of the fitted modes therefore does not
imply near-identity of the complete fields. Section~\ref{sec:line-j7-results}
reports the $J_7$ controls for these endpoints.

\subsection{Linkage does not yield a construction-independent component}
\label{sec:principal-linkage}

The common cloud also links both independent starts to the principal
warm-start teacher. Their complete residuals are negatively correlated
(Table~\ref{tab:principal-linkage}), and their cubic fitted modes agree
poorly as functions despite large absolute coefficient cosines in the
nonorthogonal monomial basis. Signed amplitude calibration does not repair
this discrepancy: principal--independent scalar transfers have held-out
$R^2=0.047$--$0.181$, versus target-specific cubic scores
$0.556$--$0.780$. The agreement of $A,B$ is therefore not common to all
three retained constructions. The principal teacher has a different
initialization lineage and refinement budget; the contrast does not isolate
initialization alone or demonstrate multiple optimal metrics.
Negative fitted-mode correlation is compatible with a positive calibrated
transfer score because the fitted amplitude may be negative. The comparison
does not show whether this discrepancy persists for a continued principal
checkpoint; continuation drift and initialization lineage remain confounded.

\begin{table}[tb]
\centering
\small
\caption{Principal teacher $P$ versus independent starts on the line-grouped
$k=10$ cloud. Intervals are conditional 95\% paired line-bootstrap intervals.
The final column compares fitted cubic functions, not coefficient vectors.}
\label{tab:principal-linkage}
\begin{tabular}{llrr}
\hline
measure & pair & residual correlation [interval] & fitted-mode correlation\\
\hline
intersection & $P,A$ & $-0.3804\;[-0.4074, -0.3538]$ & $-0.4797$\\
intersection & $P,B$ & $-0.3695\;[-0.3934, -0.3456]$ & $-0.4522$\\
volume weighted & $P,A$ & $-0.2312\;[-0.2548, -0.2093]$ & $-0.3379$\\
volume weighted & $P,B$ & $-0.2247\;[-0.2483, -0.2024]$ & $-0.3037$\\
\hline
\end{tabular}
\end{table}

\subsection{Line-grouped phase-sensitive controls}
\label{sec:line-j7-results}

The $J_7$ extension improves held-out prediction for all three $k=10$
teachers beyond both specified modulus-only libraries
(Table~\ref{tab:line-j7}). All six conditional line-bootstrap intervals
exclude zero for each library. For the principal teacher, quartic
$R^2$ increases from 0.8858 to 0.9280 under intersection sampling and
from 0.8570 to 0.9011 under volume weighting. Increasing the modulus-only
degree from three to four therefore does not remove this particular gain.
These scores use the separate three-teacher linkage cloud, not the
seven-teacher degree/replication cloud of
Tables~\ref{tab:j7} and~\ref{tab:degree-quartic-j7}; the principal checkpoint
is unchanged, but its fitted scores need not coincide across sampled clouds.

The quartic weighted designs are poorly conditioned: their condition
numbers are about $6.6\times10^{12}$--$1.1\times10^{13}$, with numerical
rank 66 or 67 of 70 at the default cutoff. This is not an exact algebraic
rank statement. The two additional cutoffs retain 62--69 directions;
all six quartic point-estimate gains remain positive. Conditional intervals
in the table use the default fit, not a cutoff selected on test scores.
These controls support predictive value beyond these two libraries,
not irreducible phase dependence, completeness of the controls, or a
universal residual correction.

\begin{table}[tb]
\centering
\caption{Adding $J_7$ at $k=10$ on the common line-disjoint split.
Baseline columns give held-out modulus-only $R^2$; increments are
held-out $\Delta R^2$ with conditional 95\% paired line-bootstrap
intervals. $P$ is the principal teacher. Values are rounded to five
decimal places to display the incremental gains.}
\label{tab:line-j7}
\scriptsize
\setlength{\tabcolsep}{3pt}
\medskip
\begin{tabular}{llrrrr}
\hline
measure & teacher & cubic $R^2$ & quartic $R^2$ & cubic increment [interval] & quartic increment [interval]\\
\hline
intersection & $P$ & 0.71831 & 0.88581 & 0.01081 [0.00819, 0.01352] & 0.04224 [0.03820, 0.04669]\\
intersection & $A$ & 0.77986 & 0.85164 & 0.00392 [0.00229, 0.00559] & 0.01010 [0.00774, 0.01248]\\
intersection & $B$ & 0.66192 & 0.71649 & 0.01070 [0.00817, 0.01305] & 0.02118 [0.01766, 0.02469]\\
volume weighted & $P$ & 0.63357 & 0.85697 & 0.00719 [0.00500, 0.00944] & 0.04414 [0.04005, 0.04806]\\
volume weighted & $A$ & 0.67248 & 0.76932 & 0.00394 [0.00184, 0.00626] & 0.01245 [0.00882, 0.01614]\\
volume weighted & $B$ & 0.55556 & 0.62505 & 0.01128 [0.00830, 0.01423] & 0.02455 [0.02015, 0.02901]\\
\hline
\end{tabular}
\end{table}

\FloatBarrier
\subsection{Cross-degree scalar transfer}

Residual agreement at fixed degree did not extend across degrees.
On the main sequence, intersection correlations were approximately
0.54 for $k=8$ versus 10, 0.56 for $k=10$ versus 12, and
0.002 for $k=6$ versus 12. Volume-weighted correlations were approximately
0.62, 0.64 and 0.067, respectively. The intersection $k=6,10$ correlation
was negative, $-0.123$ with conditional line-bootstrap interval
$[-0.208,-0.041]$.
The retained fields therefore do not behave as scalar multiples of a
common function. Their fitted cubic modes also lost accuracy when
transferred with only an amplitude adjustment.

\begin{figure}[tb]
\centering
\includegraphics[width=0.78\linewidth]{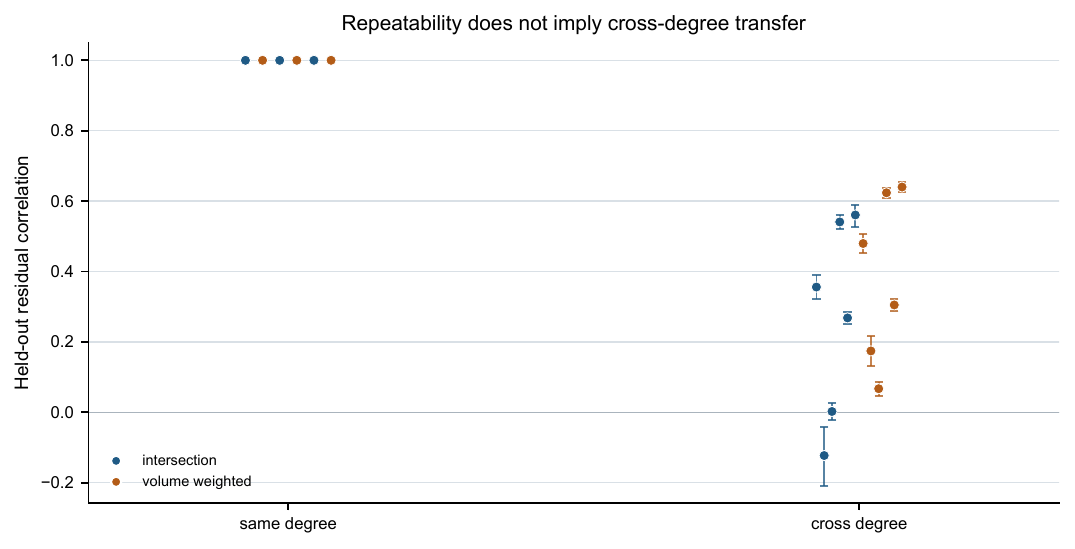}
\caption{Held-out residual correlations for warm-start fixed-degree replication and
cross-degree comparison. Points show intersection and volume-weighted
comparisons on the shared 50,000-point cloud. Intervals are conditional
95\% paired line-bootstrap intervals on 2,000 held-out lines.}
\label{fig:same-cross-correlation}
\end{figure}

The warm-start same-degree and cross-degree correlations are compared directly in
Fig.~\ref{fig:same-cross-correlation}. The separation is a statement about
the tested residual fields: repeated optimization at a fixed degree is highly
repeatable within that protocol, whereas reuse across degrees is not
comparably successful. The separate independent-start comparison is given
in Table~\ref{tab:matched-independent-correlation}; it is not included in
that figure.

The residual norms are non-monotone, increasing from $k=10$ to 12.
Four computed checkpoints do not establish an asymptotic rate.

\subsection{Joint compression and transfer from other degrees}

The raw joint cubic fit placed 0.835 and 0.817 of its training energy
in the first singular direction under the two measures, largely reflecting the greater residual
amplitude at $k=6$. Unit-training-RMS normalization reduced that fraction
to 0.467 under intersection sampling and 0.575 under volume weighting.
Ranks two and three retained approximately 0.787 and 0.938 of normalized
fitted energy under intersection sampling, and 0.832 and 0.947 under
volume weighting. These measure joint compressibility of fitted fields,
not transfer to a withheld degree.

The full span of the three other fitted fields gave substantially poorer
held-out predictions than fitting the target polynomial directly
(Table~\ref{tab:transfer}). At $k=12$, cubic transfer scores were 0.188 and 0.230
against target-specific scores 0.711 and 0.624.
Raw and normalized rank-three transfer agreed as predicted by
Proposition~\ref{prop:span}. Normalization changes truncated leading
directions but cannot supply a direction missing from the complete
source span.

\begin{table}[tb]
\centering
\caption{Rank-three leave-one-teacher-out transfer versus the target's
own fit on the main common cloud. Source bases use the other three
degrees; amplitudes use target training points. Raw and normalized
full-source-span results coincide. The quartic block is a local post-hoc
sensitivity check on the same split, without additional bootstrap intervals.
Entries are held-out residual $R^2$.}
\label{tab:transfer}
\begin{tabular}{crrrr}
\hline
 & \multicolumn{2}{c}{intersection} & \multicolumn{2}{c}{volume weighted}\\
target $k$ & transfer & target fit & transfer & target fit\\
\hline
\multicolumn{5}{l}{Cubic library}\\
6 & 0.284584 & 0.958455 & 0.247642 & 0.921715\\
8 & 0.460289 & 0.893214 & 0.525306 & 0.865281\\
10 & 0.407344 & 0.729694 & 0.421562 & 0.653304\\
12 & 0.188180 & 0.710787 & 0.229902 & 0.624140\\
\hline
\multicolumn{5}{l}{Quartic library (post hoc)}\\
6 & 0.272661 & 0.965829 & 0.247400 & 0.933604\\
8 & 0.469131 & 0.931000 & 0.524195 & 0.903110\\
10 & 0.530703 & 0.893332 & 0.593254 & 0.868004\\
12 & 0.357813 & 0.978915 & 0.487140 & 0.974154\\
\hline
\end{tabular}
\end{table}

The post-hoc quartic check improves some transfer scores but leaves a gap.
At $k=12$, quartic transfer scores are 0.358 and 0.487, versus
target-specific scores 0.979 and 0.974. Increasing the library from cubic
to quartic is therefore insufficient to match the target-specific fits.
Other feature libraries or low-rank constructions may transfer better;
these tests also leave open shared asymptotic structure at better-converged
or larger degrees.

\subsection{Fixed-space augmentation and capacity-matched controls}
\label{sec:fixed-space-controls}

Holding the numerical baseline space fixed preserves the main $J_7$
findings. For the two warm-start $k=10$ teachers, quartic-baseline
$\Delta R^2$ ranges across the three cutoffs are $0.03750$--$0.03830$
under intersection sampling and $0.03913$--$0.03969$ under volume weighting;
all corresponding conditional intervals exclude zero. For the three
linkage teachers, the respective ranges are $0.01006$--$0.04230$ and
$0.01237$--$0.04422$, again with positive interval lower bounds throughout.
At $k=12$, all twelve quartic intervals contain zero: two replicas,
two measures and three cutoffs. Thus the positive $k=10$ increments and
unresolved $k=12$ increments persist without allowing the candidate to
change the retained modulus-only baseline space. These checks concern
predictive augmentation, not proof of conditional phase necessity.

The target-specific fits have many more directions than the three-source
span, so their accuracy gap alone does not determine how informative the
source span is at fixed capacity. We therefore compare it with 1,000
random spans of the same effective rank in the centered retained library.
Gaussian QR produces Haar-distributed subspaces in the weighted training
function norm. Each draw projects the target's training coordinates onto
that span and is evaluated on the same held-out lines. No draw is selected
using test performance. All source spans have effective rank three;
same-degree replicas are excluded from their construction.

\begin{table}[tb]
\centering
\small
\caption{Capacity-matched source-span controls at the default cutoff,
reported separately for each principal target degree. Each reference entry
is the 97.5th percentile of 1,000 random rank-three span scores for that
target, not a confidence bound. All scores are held-out $R^2$.}
\label{tab:capacity-controls}
\begin{tabular}{lrrrrr}
\hline
 & & \multicolumn{2}{c}{intersection} & \multicolumn{2}{c}{volume weighted}\\
library & target $k$ & source & reference & source & reference\\
\hline
cubic & 6 & 0.284584 & 0.258403 & 0.247642 & 0.236810\\
cubic & 8 & 0.460289 & 0.235909 & 0.525306 & 0.223270\\
cubic & 10 & 0.407344 & 0.198709 & 0.421562 & 0.166863\\
cubic & 12 & 0.188180 & 0.173060 & 0.229902 & 0.158067\\
\hline
quartic & 6 & 0.272661 & 0.132967 & 0.247400 & 0.133531\\
quartic & 8 & 0.469131 & 0.128156 & 0.524195 & 0.133620\\
quartic & 10 & 0.530703 & 0.125954 & 0.593254 & 0.124964\\
quartic & 12 & 0.357813 & 0.138883 & 0.487140 & 0.132533\\
\hline
\end{tabular}
\end{table}

Every source-span score exceeds its own target's random-span 97.5th
percentile, under both measures, both libraries and all three cutoffs
(48 comparisons). The result supports nonrandom shared structure within
these libraries despite incomplete cross-degree transfer. The reference
distribution depends on the chosen library and training norm; it is not a
geometric null model or a calibrated significance test. It neither removes
the gap to the larger target-specific fits nor establishes a universal
degree-independent correction.

\section{Analysis and Discussion}

\subsection{What the learning system learns}

Much of the simplicity of the uncorrected determinant ratio follows from
the hypersurface and its holomorphic volume form. A compact scalar fit
can recover this dependence without reconstructing the Ricci-flat metric.
Here the exact relation is an analytic input, not a learning result.
Subtracting it leaves the algebraic volume-form error in
Eq.~\eqref{eq:log-volume-error} as the regression target.

The invariant coefficient matrix imposes a geometric prior on the
optimized finite-degree metric. Regression then fits its residual within
a selected library. For the matched-quality pair, the fitted cubic
functions align more closely than the complete residuals and retain
cross-start predictive accuracy. They transfer poorly to the principal
warm-start construction at the same degree. The common library,
parameterization and optimizer may contribute to the pair's agreement;
it does not identify a representation-independent component or a
correction intrinsic to the Ricci-flat limit. Transfer across degrees is
also substantially weaker on the retained principal sequence.

An individual cubic fit predicts part of the error without identifying
its cause. Write
\[
E_{k,s}=\overline E_k+\varepsilon_{k,s},
\]
where $\overline E_k$ is a protocol-dependent mean and
$\varepsilon_{k,s}$ denotes variation between refinement streams.
The small differences within the related warm-start cohort constrain
variation only within that cohort. The matched-quality pair shows that
larger field differences can coexist with transferable fitted structure.
The protocol mean can contain both finite-representation error and
systematic optimization bias. A shared warm start or optimizer plateau
can reproduce the same bias, which neither small sigma nor a near-one
correlation rules out. Initialization and training randomness both vary
between the independent starts, so their effects are not causally separated.

\subsection{An orthogonal decomposition of learnable error}

Symmetry and feature selection give a second, exact decomposition.
Let $\mu$ be a $G$-invariant probability measure on $X_0$, and work in
the real Hilbert space $L^2(X_0,\mu)$ with norm $\|\cdot\|_\mu$.
For a scalar residual $E$, define its group average by
\[
P_GE(z)=\frac1{|G|}\sum_{g\in G}E(gz).
\]
Let $V$ be a finite-dimensional subspace of $G$-invariant functions,
such as the cubic power-sum space used here, and let $\Pi_V$ be its
orthogonal projector.

\begin{proposition}[symmetry and feature-space decomposition]
\label{prop:error-decomposition}
For every $E\in L^2(X_0,\mu)$,
\begin{equation}
\|E\|_\mu^2
=\|E-P_GE\|_\mu^2
+\|P_GE-\Pi_VE\|_\mu^2
+\|\Pi_VE\|_\mu^2.
\label{eq:error-decomposition}
\end{equation}
Moreover, for any $f\in V$,
\[
\|E-f\|_\mu^2
=\|E-P_GE\|_\mu^2
+\|P_GE-\Pi_VE\|_\mu^2
+\|\Pi_VE-f\|_\mu^2.
\]
\end{proposition}

\begin{proof}
Invariance of $\mu$ makes composition by each group element unitary.
The group average is idempotent, and averaging inverses shows it is
self-adjoint. Hence $P_G$ is the orthogonal projector onto the
$G$-invariant subspace. Since $V$ lies in that subspace,
$\Pi_VP_G=\Pi_V$. The three terms in
$E=(E-P_GE)+(P_GE-\Pi_VE)+\Pi_VE$ are pairwise orthogonal.
Pythagoras gives the first identity. Replacing the final term by
$\Pi_VE-f\in V$ gives the second.
\end{proof}

This is a standard Hilbert-space projection argument applied to the
geometric learning target, not a new general approximation theorem.
Its terms distinguish symmetry violation, invariant error outside the
feature space, and captured invariant structure. For an exactly invariant
teacher, $P_GE=E$: the first term vanishes, but the second need not.
It can contain higher-degree modulus dependence or phase-sensitive invariant
dependence. Both population measures are $G$-invariant: the group acts by
unitary projective transformations, preserving the induced Fubini--Study
metric and the uniform projective-line distribution; it also preserves
$p_4$ and the Calabi--Yau volume measure. Thus exact teacher invariance
implies $E-P_GE=0$ pointwise, even on a cloud not closed under $G$.
The libraries generated by $p_j$ and the real $J_7$ are also even under
complex conjugation. No conjugation constraint was imposed on $H$.
If a teacher has a conjugation-odd residual component, it lies in the
invariant complement and is orthogonal to these even libraries under
either conjugation-invariant population measure. Such a component was
not measured here and is not inferred merely from allowing complex $H$.
The $J_7$ audit compares one specified phase-sensitive feature
against the cubic modulus-only baseline at fixed degree, but does not
isolate conditional phase information or
resolve the remaining invariant complement or rank the two explanations
globally. Likewise, the matched-quality transfer result does not identify
the invariant complement as pure optimizer noise.

The identities concern population norms and exact projections. A finite
random cloud is not generally closed under $G$, and a training fit is not
the population projector. The reported held-out errors therefore do not
numerically certify all three terms. No additional decomposition
experiment is claimed here. Finally, scalar averaging $P_GE_H$ differs
from constructing $E_{H(B)}$: the metric determinant and logarithm are
nonlinear in $H$. Proposition~\ref{prop:error-decomposition} supplies no
monotonicity guarantee for $\sigma$ under matrix averaging.

\subsection{Why compression does not imply a universal correction}

A rank-three joint fit to four fields reduces dimension only modestly,
and its basis uses every target's training values. High cumulative singular
energy therefore says little about prediction from other degrees alone.
The transfer tests measure that prediction directly, allowing new
amplitudes fitted to target training data.
The capacity-matched controls in Section~\ref{sec:fixed-space-controls}
show that the source spans nevertheless carry more predictive information
than random spans of the same rank. Incomplete transfer should therefore
not be read as absence of shared structure.

For an exact relation $E_k=a_kF$, nonzero centered residuals would have
absolute correlation one. The cross-degree correlations do not support
this description of the retained checkpoints.
For $E_k=k^{-\alpha}F+o(k^{-\alpha})$, however, the prediction concerns
sufficiently large $k$ with controlled numerical errors.
The present non-monotone sequence cannot decide that limiting question.
A common refinement budget does not match initialization quality or
guarantee convergence.

The dependence of the $J_7$ gain on the modulus-only library separates
predictive value from phase necessity. A phase-sensitive feature can improve
an incomplete fit even when the target depends only on moduli.
Conversely, its unresolved increment at $k=12$ does not
exclude other phase-sensitive dependence. The quartic mode-space check
also leaves a transfer gap despite capturing more target-specific variance.
Neither augmentation establishes a universal correction or an
impossibility result for symbolic approximation. The remaining variation
can include omitted features, representation error and optimization bias.

\subsection{An open question about degree-dependent shapes}

The unresolved optimization error limits the question we can ask about
degree dependence. For each degree, suppose a fixed optimization
problem and a specified solution-selection rule define a nonzero,
population-centered reference residual $E_k^*$. The objective,
normalization and rule must be fixed in advance; a common stopping budget
does not define an optimization-independent reference.
For two nonzero fields, define their sign-insensitive shape distance by
\begin{equation}
d_\mu(E,F)=\min_{\epsilon\in\{-1,1\}}
\left\|\frac{E}{\|E\|_\mu}
-\epsilon\frac{F}{\|F\|_\mu}\right\|_\mu.
\label{eq:shape-distance}
\end{equation}
For centered fields this equals
$\sqrt{2-2|\operatorname{corr}_\mu(E,F)|}$ and vanishes exactly when
the fields are scalar multiples.

\paragraph{Open question.}
For the tested finite degrees, do nonzero distances
$d_\mu(E_k^*,E_\ell^*)$ persist once within-degree optimization error
is independently controlled, rather than merely made repeatable under
one refinement protocol?

A quantitative criterion clarifies the missing control. If computed,
population-centered fields $\widehat E_j$ satisfy independently justified
bounds $\|\widehat E_j-E_j^*\|_\mu\leq\varepsilon_j$, then
\[
d_\mu(E_k^*,E_\ell^*)
\geq d_\mu(\widehat E_k,\widehat E_\ell)
-\frac{2\varepsilon_k}{\|\widehat E_k\|_\mu}
-\frac{2\varepsilon_\ell}{\|\widehat E_\ell\|_\mu}.
\]
Indeed, normalization changes a nonzero field by at most twice its
absolute perturbation divided by its computed norm, and the triangle
inequality applies after either sign choice. A positive lower bound would
exclude a common scalar shape for those reference residuals.
All fields entering the normalization must be nonzero;
$\varepsilon_j<\|\widehat E_j\|_\mu$ is sufficient to ensure this for
$E_j^*$. Positivity of the lower bound requires the sum of the two
relative-error terms to be smaller than the observed shape distance.
If residual norms decrease with degree, fixed absolute optimization-error
bounds become less informative. Training-mean-centered empirical fields
are not exactly population-centered: their centering discrepancy must
be included in $\varepsilon_j$, or removed by population centering before
using the correlation identity.
Repeated refinement seeds do not provide the required bounds; nor do
line-bootstrap intervals control optimization error. Finite-cloud
estimation error would also need to be accounted for when using this
criterion. Thus the question remains open, and its resolution at these
degrees would still not decide a large-$k$ asymptotic expansion.

\subsection{Scope and reuse}

The compact positive parameterization enforces finite symmetry at every
evaluation. These controls complement balanced-metric and variational
constructions \cite{Donaldson2009,HeadrickNassar2013} and help interpret
symbolic fits alongside symmetry-guided approximation
\cite{MirjanicMishra2025}. The determinant identity, comparison at common
points and source-span test do not depend on this particular optimizer.

The evidence is restricted to the Fermat point, four degrees, one scalar
observable and the tested algebraic-metric constructions. It does not
measure curvature accuracy or full-metric convergence. Two independent
$k=10$ starts do not characterize a population of initialization outcomes,
and no $k=6$ replication was performed. Line bootstraps quantify evaluation
uncertainty, not shared implementation error or optimizer bias.

Interpreting a residual as an intrinsic finite-degree correction would
require independent control of optimization and representation errors.
Initialization strategy and training budget remain confounded in the
principal-to-independent comparison. Distinct solver families and
alternative algebraic teachers, including positive tensor-network
parameterizations \cite{Wang2026TensorNetwork}, offer further tests of this
distinction. Additional degrees or invariant candidates alone would not
resolve it.

\section{Conclusion}

After subtracting the known Fermat determinant reference, polynomial
regression finds repeatable structure in the volume-form errors of the
tested warm-start metrics. Late-stage continuation changes are smaller
than the cross-degree discrepancy. The matched-quality independent pair
has less similar complete residuals but closely aligned cubic fits that
retain cross-start predictive accuracy. It satisfies the transfer
criterion and fails the full-field agreement criterion. Its agreement
does not extend to the principal warm-start construction. Across degrees,
scalar and source-derived cubic modes fall short of target-specific fits
on line-disjoint evaluations. The quartic source-span check leaves a gap
as well.

Adding $J_7$ improves the cubic fits for all seven warm-start teachers.
With a quartic modulus-only baseline, its gain remains positive at
$k=6,8,10$ but is unresolved at $k=12$. These results establish neither
phase necessity nor phase independence, and the added feature does not
resolve cross-degree transfer.

The fitted structure is stable within some tested protocols, but the
comparisons do not establish a unique complete residual or a universal
geometric correction. Replication and transfer are not certificates of
convergence. The reported intervals account for parent-line dependence
and remain conditional on the retained teachers and fitted models.

\section{Data and Code Availability}

The code, selected metric checkpoints, evaluation clouds, numerical results
and reproducibility records are available on Zenodo \cite{Eng2026Zenodo}.
The archive includes instructions for replaying the stored-cloud analyses and
documents the limits of the reproduction checks. Reproducibility metadata
are summarized in Appendix~\ref{app:reproducibility}.

\section{Declaration of Generative AI Use}

OpenAI ChatGPT and Codex assisted with programming and language editing.
They also supported the mathematical exposition and manuscript organization.
The author is responsible for checking derivations, reproducing the
analyses and approving the final interpretation before submission.

\appendix
\clearpage

\section{Reproducibility details}
\label{app:reproducibility}

\subsection{Random-stream identifiers}

These identifiers specify the pseudorandom streams used in the reported
experiments. They distinguish reproducibility metadata from the experimental
roles used in the main text. Starts $A$ and $B$ denote the two long-budget
independent runs throughout.

\begin{table}[H]
\centering
\small
\setstretch{1.0}
\caption{Optimization stream identifiers. Replication varies stochastic
refinement within the stated initialization protocol, not solver family.}
\begin{tabular}{lr}
\hline
role & identifier\\
\hline
principal warm-start sequence & 20260916\\
warm-start replication at $k=8,10$ & 314159\\
warm-start replication at $k=12$ & 20260921\\
30,000-update independent-start control & 20261011\\
continuation control & 20261001\\
90,000-update independent start $A$ & 20261031\\
90,000-update independent start $B$ & 20261032\\
\hline
\end{tabular}
\end{table}

\begin{table}[H]
\centering
\small
\setstretch{1.0}
\caption{Sampling and train/test-split identifiers explicitly specified
for the main comparisons. The paired quality check evaluates complete
clouds and therefore has no regression split.}
\begin{tabular}{lrr}
\hline
comparison & sampling & split\\
\hline
30,000-update independent-start comparison & 20261014 & 20261015\\
long-budget paired quality check & 20261101 & ---\\
line-grouped principal/independent linkage & 20261201 & 20261202\\
line-grouped degree/replication evaluation & 20261211 & 20261212\\
\hline
\end{tabular}
\end{table}

\subsection{Numerical validation}

The main methods specify the scientifically relevant sample sizes,
train/test separation, objective and optimization budgets. In the code,
the importance-weight formula includes fixed factorial and $\pi$
normalizations; global factors cancel from the normalized statistics.
The implemented negative-$\eta$ penalty is inactive for positive algebraic
metrics in exact arithmetic. Both completed line-grouped clouds retained
all attempted five-root lines; this
does not bound rejection rates in other settings.
Interrupted long-budget runs restored the parameters, Adam state and
training RNG without resetting the cosine schedule.

The two line-grouped clouds were checked for complete five-root groups,
disjoint training and test lines, homogeneous normalization and agreement
of stored power sums and $J_7$ with the coordinates. Recomputing the
regressions from fixed determinant targets in a second NumPy environment
agreed to less than $10^{-8}$ in reported $R^2$-related quantities.
This validates the regression calculation from stored targets, not metric
evaluation from checkpoint matrices or independent reproduction of training.

\subsection{Complete cubic candidate-augmentation results}
\label{app:complete-j7}

Table~\ref{tab:complete-j7} gives all seven warm-start teachers under
both measures on the degree/replication cloud. These are the original
joint-design cubic augmentations summarized in Section~\ref{sec:j7-method},
not the separate fixed-space control. Intervals condition on the retained
teachers and training fits and are not simultaneous across comparisons.

\begin{table}[H]
\centering
\small
\setlength{\tabcolsep}{4pt}
\caption{Complete cubic $J_7$ augmentation on the seven-teacher cloud.
$P$ denotes a principal teacher and $R$ its warm-start replica.
Baseline and augmented scores are held-out $R^2$; intervals are conditional
95\% paired test-line bootstrap intervals for the increment.}
\label{tab:complete-j7}
\begin{tabular}{lrlrrl}
\hline
measure & $k$ & role & baseline & augmented & $\Delta R^2$ [interval]\\
\hline
intersection & 6 & $P$ & 0.958455 & 0.991222 & 0.032767 [0.028133, 0.037929]\\
intersection & 8 & $P$ & 0.893214 & 0.897770 & 0.004556 [0.003471, 0.005569]\\
intersection & 8 & $R$ & 0.892466 & 0.897004 & 0.004539 [0.003536, 0.005484]\\
intersection & 10 & $P$ & 0.729694 & 0.738403 & 0.008710 [0.006529, 0.010962]\\
intersection & 10 & $R$ & 0.728297 & 0.737096 & 0.008798 [0.006633, 0.011155]\\
intersection & 12 & $P$ & 0.710787 & 0.717641 & 0.006854 [0.004726, 0.008929]\\
intersection & 12 & $R$ & 0.709913 & 0.717195 & 0.007281 [0.005079, 0.009616]\\
\hline
volume weighted & 6 & $P$ & 0.921715 & 0.984511 & 0.062795 [0.058452, 0.067534]\\
volume weighted & 8 & $P$ & 0.865281 & 0.874181 & 0.008900 [0.007282, 0.010482]\\
volume weighted & 8 & $R$ & 0.864576 & 0.873448 & 0.008872 [0.007308, 0.010467]\\
volume weighted & 10 & $P$ & 0.653304 & 0.658447 & 0.005142 [0.003101, 0.007282]\\
volume weighted & 10 & $R$ & 0.652517 & 0.657709 & 0.005192 [0.003165, 0.007191]\\
volume weighted & 12 & $P$ & 0.624140 & 0.634984 & 0.010844 [0.007942, 0.013691]\\
volume weighted & 12 & $R$ & 0.622942 & 0.634454 & 0.011513 [0.008274, 0.014468]\\
\hline
\end{tabular}
\end{table}

\subsection{Initialization sensitivity at a fixed refinement budget}
\label{app:initialization-control}

A separate $k=10$ control used the principal teacher's objective, batch
sizes, learning-rate schedule and 30,000-update refinement budget, but
started from an independently perturbed invariant identity. Its final
validation sigma was $0.323\%$, versus $0.0436\%$ for the principal teacher.
Residual norms and scalar transfers are summarized in
Table~\ref{tab:independent-start}. The two teachers differ substantially
in accuracy, so this comparison cannot establish initialization dependence
among equally accurate metrics.

\begin{table}[H]
\centering
\scriptsize
\setlength{\tabcolsep}{3pt}
\caption{Accuracy-unmatched initialization control on 50,000 common points.
Scores use an 80:20 individual-point split, not line-disjoint holdout, and
are descriptive without line-cluster confidence intervals. $P$ and $I$
denote the principal and independent teachers; transfer scores are $R^2$
after one target-training amplitude calibration.}
\label{tab:independent-start}
\begin{tabular}{lrrrrr}
\hline
measure & RMSE$_P$ & RMSE$_I$ & correlation
 & transfer $P\!\to\!I$ & transfer $I\!\to\!P$\\
\hline
intersection & 0.000661 & 0.004717 & -0.085216 & -0.000471 & -0.000648\\
volume weighted & 0.000561 & 0.004206 & -0.081823 & 0.004485 & 0.002669\\
\hline
\end{tabular}
\end{table}

\end{document}